%% file: main.tex
\documentclass[10pt,twocolumn]{article} 
\usepackage{latex8}

\usepackage{epsfig}
\usepackage{multirow}

\usepackage{amsmath}
\usepackage{amssymb}
\usepackage{graphicx}
\usepackage{times}

\def\mvp{\vspace*{-0.1in}}

\def\mvp{\vspace*{-0.1in}}

\begin{document}


\title{A Measure for Dialog Complexity and its Application in Streamlining Service Operations}


             
\author{Q. Vera Liao, Biplav Srivastava, Pavan Kapanipathi
 \\
            IBM T J Watson Research Center}

\maketitle

\input{abstract}


\input{introduction}
\input{background}

\input{sol-complexity}
\input{sys-arch}

\input{experiments}
\input{usage}

\input{conclusion}

\bibliographystyle{latex8}


\bibliography{complexity}

\end{document}

%% file: abstract.tex
\begin{abstract}

Dialog is a natural modality for interaction between 
customers and businesses in the service industry. 
As customers call up the service provider, their
interactions may be routine or extraordinary. 
We believe that these interactions, when seen as dialogs, 
can be analyzed to obtain a better understanding of customer needs and how to efficiently address them. 
We introduce the idea of a dialog complexity measure to characterize multi-party interactions, propose a general data-driven method to calculate it, use it to discover insights in public and enterprise dialog datasets, and demonstrate its beneficial usage in facilitating better handling of customer requests and evaluating service agents. 
\end{abstract}
   \vspace*{-2mm}

%% file: introduction.tex
   \vspace*{-6mm}
\section{Introduction}
\label{sec:intro}
\mvp

Service industry thrives on engaging customers using a company's offerings, and dialogs, whether written or spoken, is a common form of such an interaction. Over time, organizations collect a sizable volume of dialog data that may be proprietary or public depending on how customer service is provided. 

As customers call up their service providers for requests, their interactions may be routine or extraordinary. Recently, there has been significant interest in the field of service management to automatically analyze such interaction data to characterize different service sessions, types of customers, and service domains. Such characterization can help understand individual customer needs and facilitate more satisfying and cost-effective service handling. For example in \cite{qart}, NLP techniques were introduced
to track high-level indicators such as sentiments as customer interactions progress in a service center to enable managers to take pro-active actions. 

Continuing on this theme, we propose a measure of \textit{dialog complexity} to characterize service interactions with customers. Specifically, we measure complexity of service dialogs at the levels of utterances, turns and overall dialogs. The method takes into consideration the concentration of domain specific terms as a reflection of customer request specificity, as well as the structure of the dialogs as a reflection of customer demand for quantity of service actions. We propose a system architecture that automates the dialog complexity calculation, including discovery of domain-specific terms, to make it highly amenable to scale-up to new domains.

Using this measure, service providers can differentiate between simple and complex service dialogs, and take the complexity feature into consideration to improve service handling and service evaluation of agents. By applying the complexity measure to historical datasets, insights can be derived on the causes and implications of varying levels of complexity. Such insights can be used to further improve service handling and customer satisfaction. For example, it would be more satisfying and also cost-effective to allocate service dialogs expected to have high complexity to agents that are more experienced but potentially expensive.

To manifest the usage of the complexity measure, we conduct multiple experiments in the paper using dialog datasets from online repositories as well as contact centers of service providers. We show that the measure can capture the large diversity in the complexity of service dialogs. Although detailed experiments are shown later, for illustration, see Table~\ref{tab:eg} where user utterances in four service domains are shown of low and high complexity by our measure. By comparing the complexity of different kinds of dialogs and across different service domains, we show that many factors can contribute to varying dialog complexity, including service contexts and speaker characteristics.

The dialog complexity measure and insights about complexity variations can have wide usage in service industry for managing customer requests, internal processes and optimizing delivery systems. We discuss these possibilities and also propose a service agents evaluation metric that takes into consideration the complexity of dialogs they handle, and show that it makes a difference from the conventional evaluation metrics.

\begin{table*}
{\scriptsize
\begin{center}
\begin{tabular}{|c|l|c|l|c|}\hline

&{High Complexity}&{Score}&{Low Complexity}&{Score}
  \\\hline\hline
Ubuntu technical support& sudo adduser user group & 1 & that's my impressions	& 0.25\\\hline
Insurance support &will homeowners insurance cover flooring?	& 1	& what are some examples of annuities?	& 0.5\\\hline
Human Resource support& are company email addresses case sensitive?	& 0.92 &where am i?	& 0.33\\\hline
Restaurant booking agent & the lucky star serves Chinese food	& 0.94 & coke it is	& 0.33\\\hline
\end{tabular}
\caption{Examples of utterances and its complexity scores in each datasets}\label{tab:eg}
\end{center}
}
\end{table*}

To summarize, this paper makes the following contribution:
 (a) introducing the notion of dialog complexity to understand and compare dialogs in a repository  
 (b) proposing a automatic method to calculate it and providing a publicly available API to calculate dialog complexity for four different service domains
 (c) using it to understand varying customer interactions in a variety of domains using public and proprietary data
 (d) demonstrating its usage to improve service management operations.
   
 In the rest of the paper, we first give a background about service dialogs, and introduce the datasets we experiment with. Next, we motivate the desirable characteristics of a dialog complexity measure and propose a method to calculate it. We then conduct experiments to show that the proposed measure can characterize diverse customer interactions, and to verify that the measure captures service request specificity and quantity. Finally, we discuss its usages in improving service dialog handling and evaluation.
 
   \vspace*{-2mm}

%% file: background.tex
\section{Related Work and Background}
\label{sec:background}

Service science~\cite{service-science} deals with principled approaches to drive innovations in the service ecosystem. For the purpose of the paper, we deal with the interaction (e.g., chat dialog) that a client  conducts with a service provider in order to resolve a problem with the product or service the client is interested in.

There is a large volume of prior works on designing, monitoring and evaluating service dialogs in fields such as marketing, management and service computing (e.g.~\cite{anton2000past,bitner1990evaluating,christopher1991relationship}). This paper is most relevant to prior work developing measures to characterize and compare service dialog sessions. While most previous literature focused on quality and satisfaction measures through customer surveys (e.g.,~\cite{jaiswal2008customer,cronin1992measuring}), recent work starts exploring text analytical methods to directly derive measures from dialog contents, such as sentiment based measures~\cite{qart}. Our focus is different in that, beyond evaluation, we are also interested in \textit{optimization} of dialog handling, and thus, we focus on measuring the complexity as a characteristic of a dialog session, instead of the outcome. 

There is a rich literature on analyzing dialogs. In social science, conversation analysis deals with identifying regular patterns in dialogs and the underlying behavioral reasoning \cite{hutchby2008conversation}. In computer science, dialog analyses are often driven by advancement of speech and dialog systems \cite{mctear2002spoken}, focusing on developing NLP and machine learning methods to understand, predict and evaluate dialogs (e.g. \cite{young2013pomdp}).

Recently, new methods are developed to \textit{profile} dialogs in different domains, or for different dialog systems, and \textit{complexity} has been proposed as a data-driven metric for such purpose \cite{open-concept,domcat,pollard-complex,rauterberg-complex}. These studies were primarily driven by informing the implementation of dialog systems, and tended to focus on assessing the size of domain entities or concepts. In linguistics, dialog complexity has been studied from human readability point of view by identifying linguistic markers for a more or less elaborate styles \cite{biber1992complexity}. In this paper, we proposes a dialog complexity measure considering multiple dimensions of dialogs to enable profiling of diverse services dialogs, and to facilitate the interpretation of complexity profiles for service handling.
\vspace*{-5mm}
\subsubsection{Scope and definition of service dialogs:} This paper focuses on {\em service dialogs}. A service provider may: 1) use a dedicated contact center where one agent responds to one user at a time; or 2) use a public forum where both agents and other users may respond. We consider both types of dialogs in this paper. The interactions may be written or spoken, and in the case of the latter, we assume to have a transcribed version of the dialog. Recently, automated agent systems, in the forms of spoken dialog system or chatbot, have been on the rise. Our measure does not differentiate between dialogs with human or automatic agents. We will experiment with datasets from both.

A {\em dialog} is made up of a series of {\em turns}, where each turn is a series of {\em utterances} by one or more participants playing one or more {\em roles}. In the example of customer support center, a user contacts a service center and enters into a dialog with a customer support agent. The participant roles here are that of a customer and an agent, and the roles inter-leave in every turn. On the other hand, in the example of online support, a person may raise an issue on a public portal and anyone may reply. The role  of all participants here is a portal user. Since questions and answers do not necessarily happen in pair, we consider each user utterance in such a case of single role to define a new turn.

\section{Service Dialog Datasets used in Experiments}
\label{sec:dataset}

We will conduct experiments with the following four dialog datasets with service agents (both human and automated agents) working in different service domains:

    \noindent{\bf Public-Ubuntu technical support:} This corpus is scraped from Ubuntu online support IRC channel, where users post questions about using Ubuntu. We obtained the original dataset from \cite{lowe2015ubuntu}, and selected 2 months of chatroom logs. We extracted `helping sessions' from the log data, where one person posted a question and other user(s) provided help. The corpus contain both dyadic and multi-party dialogs.
    
    \noindent{\bf Public-Insurance QA:} This corpus contains questions from insurance customers and answers provided by insurance professionals. The conversations are in strict Question-Answer (QA) format (with one turn). The corpus is publicly available provided by \cite{feng2015applying}.

    \noindent{\bf Public-Restaurant reservation support:} This corpus contains conversations between human users and a simulated automated agent that helps users find restaurants and make reservations. The corpus was released for Dialog State Tracking Challenge 2 \cite{henderson2014second}.

    \noindent{\bf Enterprise-Human Resource bot:} This corpus is collected from internal deployment of an HR bot - a virtual assistant on an instant messenger tool that provides support for new hires. Although the bot does not engage in continuous conversations (i.e. carrying memory), it is designed to carry out more natural interactions beyond question-and-answer. For example, it can actively engage users in some social {\em small talks}.  
    


\begin{figure}[h]
  \centering
    \includegraphics[width=0.4\textwidth]{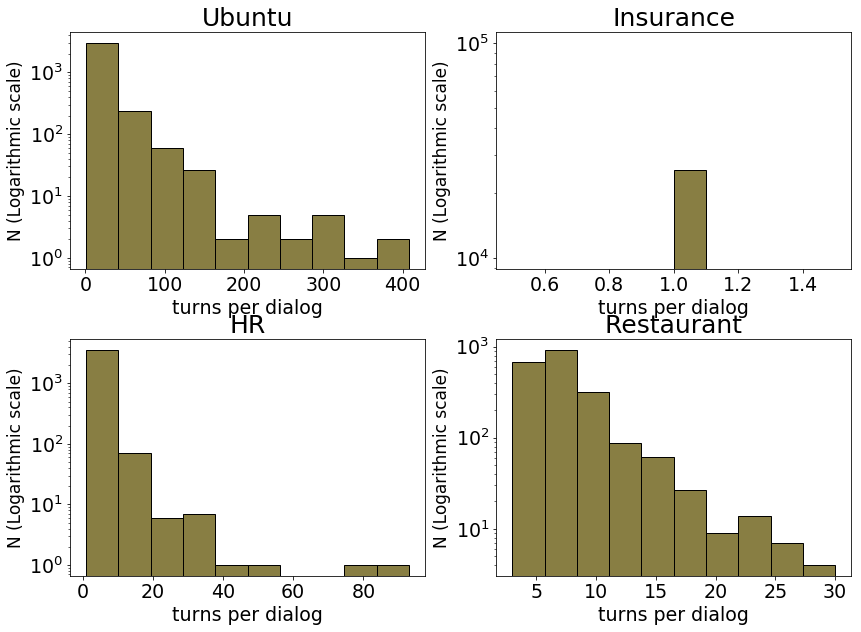}
  \caption{Distribution of number of turns per dialog}
  \label{fig:turns}
\end{figure}

In Table~\ref{tab:descriptive} (left), we present descriptive statistics of these corpora. In Figure~\ref{fig:turns}, we plot distributions of number of turns per dialog.  Except for Ubunutu, we define turn as a series of utterances where both the customer and the agent (i.e., all roles) finished speaking in one round. In Ubuntu's open, multi-party, IRC context, all participants have the same role, and hence, we define turn to be the same as utterance.
From the table and plots, we can observe several characteristics of dialogs varying across these service domains. For example, Ubuntu IRC tend to be long dialogs. Insurance QAs are strictly in one turn. Dialogs with the HR bot tend to be short with large variations, while Restaurant booking ones have less variations in length. 


\begin{table*}
\begin{center}
{\scriptsize  
\begin{tabular}{|l|c|c|c|c||c|c|c|c|}\hline

&{N (dialog)}&{M (turns/ dialog)}&{M (utt./ turn}&{M (words/ utt.)}&{Specialized domain* }& {Standardized procedure}&{QA}
  \\\hline\hline
Ubuntu& 3318& 18.53&1&17.90&0.11&&\\\hline
Insurance&25499&1&2&17.02&0.22&\checkmark&\checkmark\\\hline
HR & 3600& 2.47&2&51.11&0.25&&\\\hline
Restaurant & 2118& 7.37&2&8.24&0.27&\checkmark&\\\hline

\end{tabular}
}
\caption{Left:Descriptive statistics of corpora; Right: Features of dialog domains ( *\% of domain words overlapping with common English words reflecting domain specificity.)} \label{tab:descriptive}

\end{center}

\end{table*}

We can postulate that dialog contexts of these corpora differ in several key dimensions (Table \ref{tab:descriptive} right): 1) Ubuntu is the most specialized domain among the four, because it involves a large number of specialized vocabularies, and the tasks are less commonplace. To verify this, we calculate the percentage of domain specific words overlapping with common English words (extraction method to be discussed in Section \ref{sec:sys-arch}). The idea is that the lower the percentage, the more specialized the domain is. As expected, we found that Ubuntu has significantly lower percentage(Table \ref{tab:descriptive} column ``specialized domain''). 2) Insurance and Restaurant booking are more customer centric with standardized processes than the other two. Insurance dialogs are strictly problem-solving question and answer between customer and agent, while Restaurant dialogs are focused interactions where the agent collects customers' preferences for reservations with a set of pre-defined criteria. There are more variations, and also more \textit{off-topic discussions} in Ubuntu and HR. 3) Only insurance dialogs are in QA format.


%% file: sol-complexity.tex

\section{Calculating Dialog Complexity}
\label{sec:complexity}

The desiderata from a dialog complexity measure are that it can: 
    (a) be automatically calculated;
    (b) be agnostic to the representation (e.g., intents, entities) and yet be able to incorporate them where available;
    (c) allow comparison of dialogs;
    (d) be easy to interpret source of complexity;
    (e) be composable over dialog structure to allow ease of computation and any relative weighing;
    (f) support boundary condition properties. 

Given our focus on service dialog handling, the boundary conditions are: (a) complexity of an utterance with less complex words should be less than or equal to the same utterance with more complex words. While other definitions are possible, we define word complexity in terms of domain specialization, as more domain specific words may reflect more demanding service dialogs for specificity and efficiency;  
(b) if utterance complexity equals, complexity of a turn with less participants should be less than or equal to a turn with more participants; (c) if content complexity equals, complexity of shorter dialogs should be less than or equal to longer dialogs in the same domain. Based on the desiderata, we define a set of complexity measure at the levels of utterance, turn and dialog.

\subsection{Complexity at Utterance Level}

Let {\em SWL} represent the set of stop words,
{\em ES} stand for the set of English subset (common words),
{\em DS} for domain specific words
and rest of the words are part of noise set {\em NS}. An utterance $U$ consists of word phrases $w_i$
such that $\mid U \mid$ = $N_U^w$ = $\sum_{1}^{\mid U \mid} w_i$. We define the complexity of a word phrase $w_i$, denoted $c(w_i)$, by following terms in the given order
\vspace*{-2mm}
\begin{equation}
    c(w_i) = \left\{
    \begin{array}{c l}	
           1 & \mid w_i \in DS   \\
         0.5 & \mid w_i \in ES   \\
           0 & \mid w_i \in SWL 
    \end{array}\right.
\end{equation}

We define the  complexity of an utterance, denoted
$c(U)$, by
\vspace*{-2mm}
\begin{equation}
    c(U) = \frac {\sum_{i=1}^{\mid U \mid} c(w_i)} {\mid U \mid}
\end{equation} 

For experiments, we apply the following:

\begin{itemize}
    \item {\em SWL}: default English stop words\footnote{At: http://www.ranks.nl/stopwords}
    \item {\em ES}: Over 2000 common English words\footnote{http://www.talkenglish.com/vocabulary/top-2000-vocabulary.aspx}
    \item {\em DS}: top $k$ word phrases of a domain
    obtained from domain specialist, frequency or other methods. In the next section we present a term frequency based method to automatically identify DS for each service domain.
\end{itemize}

\subsection{Complexity at Turn Level}

A turn is a collection of utterances where each role gets to speak at least once. For a 2-role dialog, a turn consists of two utterances. We propose two definitions of turn complexity. The first one is averaging utterance complexity within a turn, calculated by the following:
\vspace*{-2mm}
\begin{equation}
c(T) = \frac {\sum_{i=1}^{ \mid T \mid } c(U_i)} { \mid T \mid }
\end{equation}

where the number of utterances $U_i$ within the turn $T$ is denoted by $\mid T \mid$. 

Since turn complexity can be seen as a way to reflect the complexity of domain specific information exchange in the turn, in another definition, we introduce dialog acts tag to calculate a weighted sum of utterance complexity. Dialog acts are tags that indicate the communicative function of the utterance \cite{stolcke2000dialogue}. For example, an utterance may intend for requesting information, providing information, or for social functions such as greetings or closing the dialog. Several recent papers introduced NLP techniques to automatically generate dialog acts tags (e.g.~\cite{stolcke2000dialogue}).

We assume a function $\alpha(U_i)$ is available to get the dialog act tag for utterance $U_i$.  
Further, for each dialog act $j$, we denote
its weight by $w^j$ (in 0-1 range).  The weighted turn complexity is calculated by:
\vspace*{-2mm}
\begin{equation}
c(T_{DA})=\frac {\sum_{i=1}^{ \mid T \mid} c(U_i) * w^{\alpha(U_i)} } { \mid T \mid }
\end{equation}

\subsection{Complexity at Dialog Level}

The utterance and turn complexity measures defined
above focus on the content of interactions.
To measure complexity at dialog level, we ensure both
the content and its structure to be considered. The underlying assumption is that \textit{the structure such as length may reflect the quantity of task specific actions, which may be orthogonal to the concentration of task specific actions} (e.g., some customers may seek to resolve one difficult problem quickly versus others with many easy requests).
Thus, we have two components in the calculation: average turn complexity to reflect the content complexity, and the length of the dialog relative to the maximum length in the dialog dataset of that kind. The latter component can be seen as reflecting the structural complexity (length) of the particular dialog relative to the maximum structural complexity (length) that the service context allows. While we use dialog length as a simple indicator, more sophisticated structural features can be introduced. One can weigh these independent
components to arrive at the total dialog complexity.

We denote the number of turns $T_i$ in the dialog $D$ by $\mid D \mid$ = $N_D^T$. Let $N_D^{T_{max}}$ be the maximum number of turns per dialog in the dataset $S$ ($D_i \in S$). Dialog complexity is then calculated as:
\vspace*{-2mm}
\begin{equation}
c(D) = w_1 * \frac {\sum_{i=1}^{ N(t)} c(T_i)} { N^T_D} + w_2 * \frac {N^T_D} { N_D^{T_{max}}}
\end{equation}

In our experiments, we give equal weight to both with  $w_1 =w_2=0.5$. The identification of optimal weights is beyond the scope of this paper, but can be achieved by building a regression model on annotated dialog complexity.

\subsubsection{Discussion on the calculation}
While dialog complexity has been studied for readability \cite{biber1992complexity}, and size of domain concepts \cite{open-concept,domcat,pollard-complex,rauterberg-complex}, we focus on features that may impact the difficulty of service dialog handling. Specifically, we consider the concentration of domain specific words as a reflection of customer demand for specificity and/or efficiency, and length of dialogs as a reflection of customer demand for quantity of service actions.
One can conceive more advanced metrics such as by including more comprehensive list of features that can predict difficulty of dialog handling (e.g., using machine learning methods), providing additional data is available conveying signals about the handling difficulty. Recent research on neural network based dialog quality measures~\cite{kannan2017adversarial} can also be adapted. However, in this paper, we focus on an explainable measure that may help gain business trust. We leave the expansion and refinement of the metrics to future work.



      




%% file: sys-arch.tex

\section{System Architecture}
\label{sec:sys-arch}

\begin{figure*}

  \centering
    \includegraphics[width=0.7\textwidth,scale=0.3]{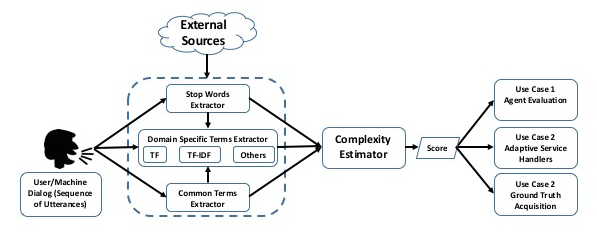}
  \caption{Architecture of the system to calculate dialog complexity}
  \label{fig:arch}
\end{figure*}
The system architecture as shown in Figure~\ref{fig:arch} is comprised of four primary modules: (1) Stop words extractor, (2) Domain specific terms extractor, (3) Common terms extractor, and (4) Complexity Calculator. While Section~\ref{sec:complexity}  described the Complexity Calculator in detail, in this section, we focus on the implementation details of the other three modules. The system is developed in Python using libraries pandas, nltk, and sci-kit learn\footnote{http://www.nltk.org/,http://pandas.pydata.org/,http://scikit-learn.org/stable/}.  The system is openly available as an API\footnote{\url{https://dialog-complexity.mybluemix.net/}}. The API, presently, has the ability to calculate complexity for dialogs in the four domains of the datasets we introduced about, i.e, HR, restaurant, insurance, and Ubuntu. Furthermore, additional domains can easily be updated with the availability of domain-specific dialog data.

As explained in Section~\ref{sec:complexity}, choice of {\em SWL}, {\em ES}
and {\em DS} play a crucial role. We experimented with a few alternatives and chose {\em SWL} and {\em ES} from online sources.
For {\em DS}, we utilize information extraction (IE) and retrieval (IR) techniques on dialog corpora.
Common IR techniques such as Term Frequency (TF) and Term Frequency-Inverse Document Frequency (TF-IDF) are used to determine the prominent, domain specific terms from the utterances~\cite{manning2008introduction}. Other techniques that perform keywords~\cite{zhang2008automatic} and key phrase extractions~\cite{witten1999kea} from documents can be plugged into the system. We fix a threshold $\delta$ to pick the top-$\delta$ percentage of domain specific terms.
Since stop words, in general, form the majority of terms in natural language documents~\cite{manning2008introduction}, we
pre-process to remove stop words before $DS$ extraction. For experiments presented in the paper, we employed TF-based {\em DS} extractor and set the threshold $\delta=50$. Note that we experimented with $\delta=20$ and $\delta=30$ , and the conclusions in Section \ref{sec:expt} hold. For the sake of simplicity, we will present the following experiments with $\delta=50$.

%% file: experiments.tex

\section{Experiments: Understanding Service Dialog Characteristics}
\label{sec:expt}

In this section, we demonstrate the usage of the complexity measure for gaining insights on the differences in service dialog interactions. Specifically, we apply the complexity measure to dialogs from different service domains (datasets introduced in Section~\ref{sec:dataset}), and different types of speakers (as an experiment, we compare customer v.s. agent), and show different complexity characteristics both at aggregate level and in procedural patterns. By interpreting the complexity characteristics, we gain insights on the contributors of dialog complexity, which can then be used to facilitate more effective handling of service requests (e.g., to accommodate the underlying needs of the more complex types of dialog).  In the experiments, we ask the following research questions:

    
    \noindent{\bf RQ1}: What complexity characteristic at aggregate level, i.e. \textit{complexity signatures}, do different service domains (i.e., datasets) have? 
    
    Given that the four datasets differ in several key dimensions (see Table \ref{tab:descriptive} right), we focus on these dimensions and ask:
    	\begin{itemize}
    	\item {\bf RQ1A}: What complexity signatures do dialogs in more specialized domain have?
    	\item {\bf RQ1B}: What complexity signatures do dialogs with standardized procedures have?
        \item {\bf RQ1C}: What complexity signatures do QA dialogs have?
    
    	\end{itemize}

    \noindent{\bf RQ2}: For multi-turn dialogs, what complexity characteristic in \textit{procedural patterns} do different service domains have?
    
    \noindent{\bf RQ3}: What complexity characteristics do different roles of speaker, specifically customer and agent, have? Does it vary for different service contexts?


\subsection{RQ1: Comparing Aggregate Complexity Across Datasets}

\begin{figure*}[h]
  \centering
    \includegraphics[width=0.31\textwidth]{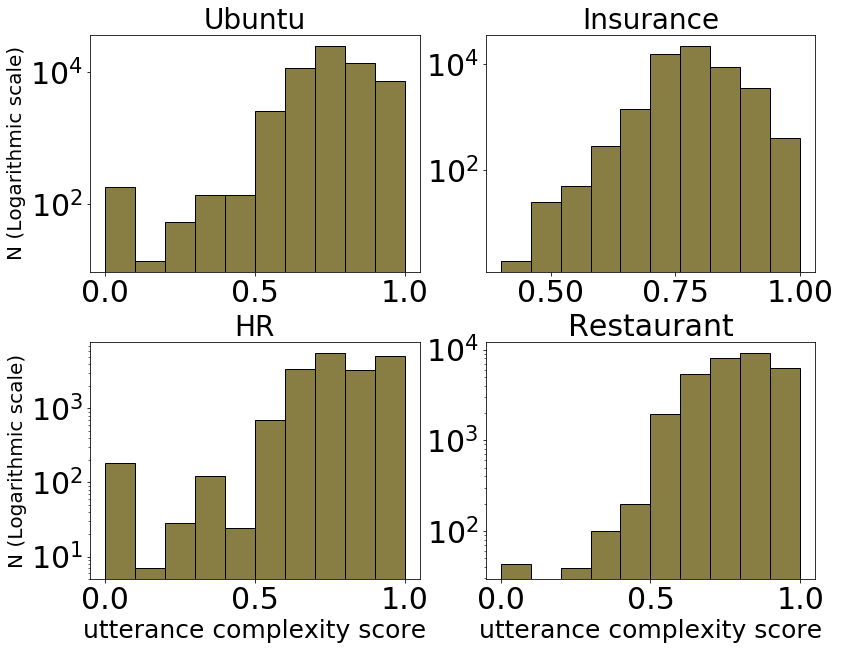}
    \includegraphics[width=0.31\textwidth]{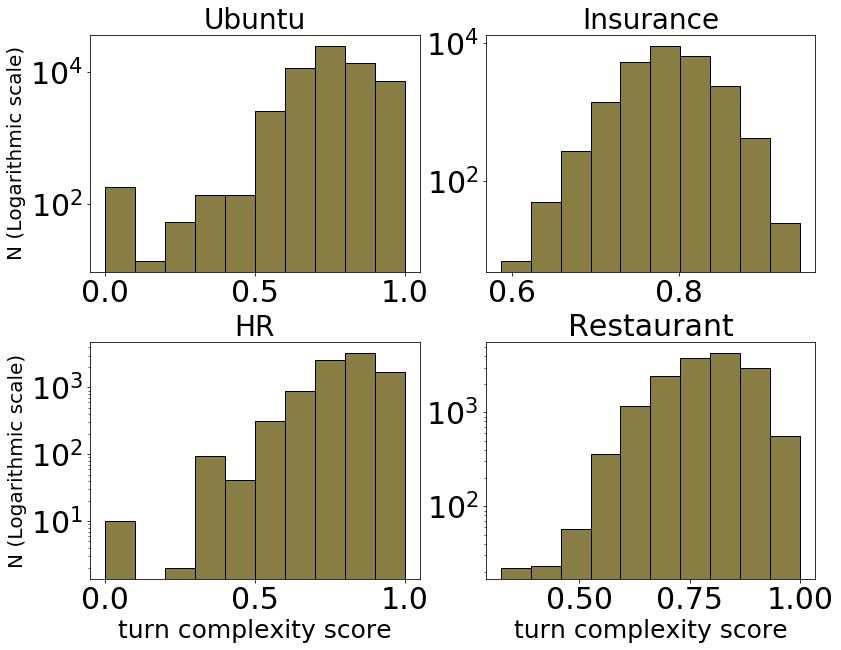}
     \includegraphics[width=0.31\textwidth]{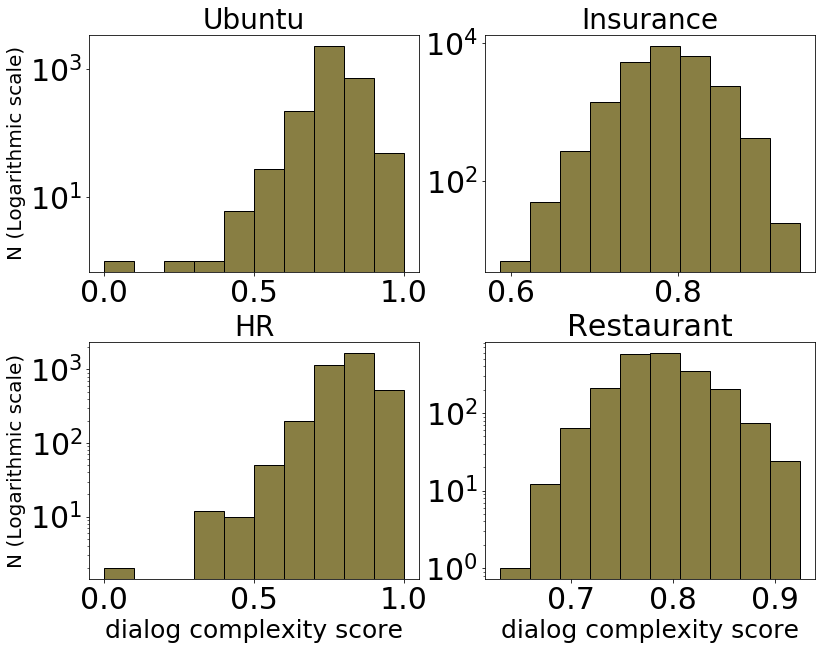}
     \vspace*{-4mm}
  \caption{Distribution of utterance(left), turn  (middle) and dialog complexity (right)}
  \label{fig:utt_cpl}
\end{figure*}

Based on the calculations specified in Section \ref{sec:complexity}, we calculate the complexity of each utterance, turn, and dialog in the four datasets. Table~\ref{tab:complexity} presents the average values of utterance, turn and dialog complexity for each dataset. Figure \ref{fig:utt_cpl} plots the distributions of utterance, turn and dialog complexity of datasets. The figures and table paint a clear picture that complexity differs across these datasets at all levels. 

\begin{table}
\begin{center}
{\scriptsize
\begin{tabular}{|l|c|c|c|}\hline

&{M (utt.)}&{M (turn)}&{M (dialog)}
  \\\hline\hline
Ubuntu&0.767&0.767&0.407\\\hline
Insurance&0.789&0.789& 0.894\\\hline
HR &0.801&0.803&0.423\\\hline
Restaurant &0.788&0.788&0.518\\\hline
\end{tabular}
}
\caption{Average complexity of each corpus}\label{tab:complexity}
\end{center}
\end{table}

Before moving on to discuss the observed differences, we first verify that the differences of complexity distribution between datasets are statistically significant. We conducted pairwise Kolmogorov-Smirnov tests for complexity score distributions at utterance , turn, and dialog levels (Figure \ref{fig:utt_cpl}). K-S test is a statistic that quantifies distances between two empirical distributions, and if significant, it means the two distributions are not the same. We found all the K-S tests to be significant (all $p<0.001$), verifying that the complexity distributions of these datasets are all different between each other.





\subsubsection{RQ1A - Domain Specialization}
\label{sec:RQ1A}

From Table \ref{tab:complexity}, it appears that dialogs in Ubuntu, the most specialized domain of the four, have the lowest average complexity at all three levels. This is reasonable since the domain-specific words extracted for Ubuntu dialogs tend to be highly specialized words, and are thus less concentrated in utterances. Put it differently, lay people are less of experts in this uncommon domain compared to the other three service domains. 

Ubuntu dialogs also have the lowest dialog complexity (Figure \ref{fig:utt_cpl} right) because the structure complexity component in Equation 5 tends to be low for most dialogs. The reason is that the maximum turn of dialog in this dataset is very high($N_D^{T_{max}}$ in Equation 5), as the open-chat environment allows free forms and flows of dialogs . 

\subsubsection{RQ1B - Standardized Service Dialog}


As noted in Section~\ref{sec:dataset}, the contexts of Insurance and Restaurant datasets are more customer centric following standardized procedure. This difference is evident in 
experiments where we note that for Insurance and Restaurant there are few occurrences of utterances and turns with low complexity (Figure \ref{fig:utt_cpl} left and middle). This is because there are far fewer off-topic discussions in these dialogs. In contrast, there is a peak in the very low end of utterance and turn complexity for Ubuntu and HR. These are either short utterance with almost only stop words, or off-topic discussions with few domain-specific words.

Another signature of standardized procedure is that the distribution of dialog complexity (Figure \ref{fig:utt_cpl} right) is more balanced. This is because the dialog lengths in these domains tend to have less variance as the requests are more consistent. Especially in the case of Insurance, users uniformly submit only one request in a session.

\subsubsection{RQ1C - QA Dialog}
\label{sec:RQ1C}
In addition to having the signatures of dialogs with standardized procedures, Insurance dialogs have a complexity signature unique to its QA nature---having the highest complexity of all datasets at all levels (all three in Figure \ref{fig:utt_cpl}). This is because all QA dialogs attempt to solve the problem within one turn. From a content point of view, all utterances have to be highly concentrated on the topic. From a structure perspective, all dialogs uniformly have the same maximum length as the particular dialog context allows only one request per session ($N_D^{T_{max}}$ in Equation 5).

\subsubsection{Turn Complexity Weighted by Dialog Act Tag}

We note that in strict sequential conversations (Insurance and Restaurant), the average turn complexity (Equation 3) of the dialog corpus would be the same as average utterance complexity, although the distributions vary. 
We now experiment with the case where dialog act is available (Restaurant dataset) for weighted turn complexity (Equation 4)

\begin{figure}[h]
  \centering
    \includegraphics[width=0.3\textwidth]{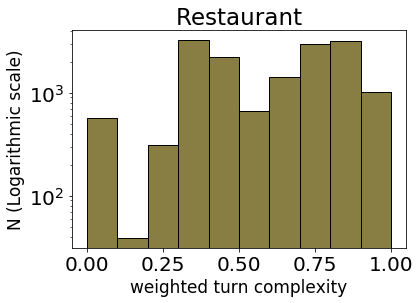}
  \caption{Distribution of turn complexity weighted by dialog act tags}
  \label{fig:w_turn_cpl}
\end{figure}

 The Restaurant dataset is published with tagging of dialog acts such as: \textit{welcome-msg}, \textit{inform}, \textit{offer}, \textit{request}, \textit{bye}, \textit{affirm}, \textit{negate}, \textit{thankyou}, \textit{confirm}, \textit{select}, \textit{acknowledge}, \textit{hello}, \textit{repeat}, \textit{deny}.
For simplicity, we only separate DA tags with social functions from the rest, and define $w^{(\textit{welcome-msg})}=w^{(\textit{bye})}=w^{(\textit{hello})}=w^{(\textit{thankyou})}=0$, all other $w^{\alpha}=1$. Figure \ref{fig:w_turn_cpl} shows the distribution of weighted turn complexity. Interestingly, the distribution becomes ``flatter'', as dialog acts tags provided additional information about the intentions of utterances that are not identifiable by simply looking at the words.

\subsection{RQ2 - Procedural Patterns of Complexity}
\mvp

In the second set of experiment, we analyze the procedural patterns of complexity, defined as how the turn complexity changes as the dialog progresses.
We use the corpora of Restaurant and Ubuntu for the experiment.

\begin{figure}[h]
  \centering
    \includegraphics[width=0.4\textwidth]{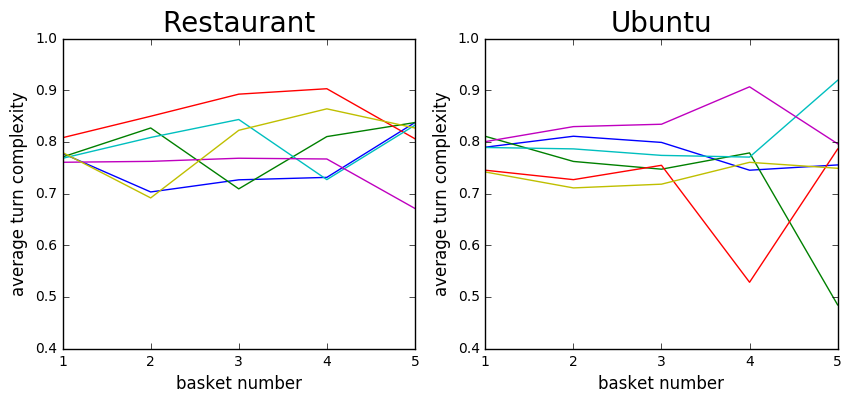}
  \caption{Kmeans clusters (K=6) of procedural patterns of turn complexity}
  \label{fig:cluster}
\end{figure}

We follow the following steps:
1) First, we eliminate dialogs with number of turns in the highest and lowest 15 percentile, leaving dialogs of 5-20 turns for HR, and 5-35 turns for Ubuntu. We evenly divide the number of turns in a fixed number of $N$ baskets. Here we set $N=5$. 2) We calculate the average turn complexity for each basket. 3) We run k-means clustering ($k=6$) to identify clusters of procedural patterns with the complexity value of the 5 baskets. 4) We use the centers of the clusters to represent the \textit{signature of procedural patterns} for the dataset, as plotted in Figure \ref{fig:cluster}.

We observe several interesting patterns: 1) For both datasets, turn complexity varies less at the beginning than at the end of dialogs, showing that they tend to have more consistent initiating patterns but diverge as the dialogs progress. 2) Dialogs in Restaurant dataset have less procedural variation than those in Ubuntu. This is consistent with the fact that restaurant booking follows a more standardized customer service procedure. 

 \subsection{RQ3 - Role Differences}
 
 In the third set of experiment, we compare the complexity characteristics of different speaker roles, i.e. customers and agents. We excluded Ubuntu since it comprises of multi-party open chat. We calculated the average utterance complexity of agent and customer for the other three datasets. The result is shown in Table \ref{tab:roles}. 
 \begin{table}[h]
\begin{center}
{\scriptsize
\begin{tabular}{|l|c|c|}\hline

&{Agent}& {Customer}
  \\\hline\hline
Insurance&0.769&0.808\\\hline
HR &0.770&0.833\\\hline
Restaurant &0.777&0.799\\\hline
\end{tabular}
}
\caption{Average utterance complexity by roles }\label{tab:roles}
\end{center}
\end{table}

Interestingly, we observed that the utterance complexity of customers is higher than agent in all three datasets. One potential explanation, we observed, is that customers tend to use more succinct phrases focusing on the requests. For example, in the dialogs with the HR bot, which were carried out by typing in a chat windows, some users used just discrete keywords such as `travel booking' instead of natural conversations. In contrast, agents tend to speak more politely, and thus using more elaborated sentences.

 \subsubsection{Discussions on the experiments}
 In the above experiments, we first show that our multi-dimension dialog complexity measure can characterize diverse service dialogs at aggregate level and by procedural patterns. By comparing the complexity signatures across different service dialog domains and speaker roles, we identify several contributors for varying dialog complexity. In Section~\ref{sec:usage}, we discuss how the dialog complexity measure can be used in combination with the insights gained from empirical analysis of historical datasets to improve handling of services. Before that, we present another set of experiments that further validate our interpretation on the contributors of complexity.
 
  
 \section{Experiments - Correlation of Complexity with Service Actions}
In defining the calculation of dialog complexity, we made two main assumptions: 1)the content based complexity (utterance and turn) should reflect user requests' domain specificity; 2) the dialog complexity should reflect the quantity of requested actions. We present two experiments below that provide validation for these assumptions.    
   
\subsection{Correlation with Domain-specific Information Retrieval Success}
To validate that utterance complexity reflects domain specificity, besides manually examining utterances with high and low complexity as in Table~\ref{tab:eg}, we choose to conduct an experiment by comparing the complexity of utterances that are successfully and unsuccessfully processed by a domain-specific information retrieval system. The idea is that utterances that can be processed by the system should be relevant to the domain, and we should expect them to have higher complexity.

Dialogs in the HR dataset are user interactions with an automatic agent using an HR Information Retrieval System (HRIRS). The knowledge base is constructed by HR knowledge about the company, with the addition of common social talks. That is, user requests unrelated to the company, such as ``find me a best Thai restaurant'', would not be successfully handled by HRIRS.


Without going to the technical details, HRIRS uses a two-level hierarchical natural classifier (NLC) to classify the \textit{intent} of an user utterance to retrieve the matching answer, which represents state-of-art dialog system technology. With the NLC, user utterances fall into three categories: 1) Correct retrieval, when the two levels of NLC are above confidence threshold and match a mapping relation. A manual evaluation of 3\% data showed that more than 87\% user utterances in this category received reasonable answers. 2) Low confidence, when either of the two levels of NLC is below confidence threshold. 3) Incorrect retrieval, when the two levels of NLC are above confidence threshold but do not match the mapping relation. The manual evaluation showed that more than 75\% in this category received wrong answers.

 \begin{table}[h]
\begin{center}
{\scriptsize
\begin{tabular}{|l|l|l|l|}\hline

&{Correct}& {Low confidence}& {Incorrect}
  \\\hline
N&4426&1142&375\\\hline
{M(complexity)}&0.786&0.725&0.637\\\hline

\end{tabular}
}
\caption{Numbers and average complexity of utterances with different HRIRS outcomes}\label{tab:retrieval}
\end{center}
\end{table}
 
Table~\ref{tab:retrieval} shows the average utterance complexity of user input in the three categories. All pair-wise t-tests are statistically significant ($p<0.001$). It indicates that user utterances that resulted in successful information retrieval tend to have higher complexity, validating our assumption that domain specificity should be reflected by the complexity measure. This may also imply differences between human-processing and machine-processing of dialogs. While casual, less domain-specialized dialogs could be easier for a human agent to handle, it may be more problematic for information retrieval with domain specific knowledge base.
  
\subsection{Correlation with Quantity of Requests}
In the next experiment, we validate whether higher dialog complexity is associated with increasing variation of requests within a service session. We use the Restaurant dataset as our case study. In a typical dialog between user and an automated agent for restaurant booking, the conversation is complicated by frequent change in customers' request, i.e. asking for a different kinds of restaurant, price range. The change may happen either because the system could not find a satisfying answer for the initial request, or because the customer changed his or her mind midway. We choose to compare the number of restaurant types in dialogs with varying complexity. 
For simplicity, narrowing down on features of a restaurant such as price range was not considered a new request variation.

We started by ranking all dialogs in the Restaurant dataset by dialog complexity in descending order. Then, we selected three groups--- high complexity (rank 1-20), median complexity (rank 1045-1064) and low complexity (rank 1999-2118) from the complexity spectrum. We manually labeled the number of restaurant type requests for all 60 dialogs, and present the average number for each group in Table~\ref{tab:request}. It shows that, the higher the dialog complexity is, the more restaurant types were in requests, validating that dialog complexity is strongly correlated ($r=0.54$) with the variations in requests. In fact, we observed that dialogs with the highest complexity are mostly ones where the users were intentionally ``breaking'' the system, by keeping asking for different kinds of restaurants and typing in repetitive, even random requests.

 \begin{table}[h]
\begin{center}
{\scriptsize
\begin{tabular}{|c|c|c|c|}\hline

&{highest complexity}& {median complexity}& {lowest complexity}
  \\\hline
M(requests)&4.20&1.45&1.05\\\hline

\end{tabular}
}
\caption{Average number of restaurant-types in different dialog complexity groups.}\label{tab:request}

\end{center}
\end{table}



%% file: usage.tex

\section{Usage of Dialog Complexity in Services}
We first discuss the usage of dialog complexity to improve service dialogs handling, using the examples and insights from the above experiments. We then explore an additional area for the usage of dialog complexity---to improve service agent evaluation.

\label{sec:usage}
\subsection{Improving Service Handling}
A direct usage of dialog complexity is to tailor service handling according to the complexity profiles of dialogs. This could be at the service context level. For example, from the above experiments, we discover that in the HR support context, users tend to speak in varying complexity, and one underlying reason is the frequent engagement in social chit-chat. Or, through comparing procedural patterns, we would expect that dialogs in Ubuntu support are less likely to follow consistent procedures compared to restaurant booking. These insights can be taken into consideration when training human agents or developing automatic agents.  

The usage could also be to tailor service handling for different types of requests or customers, potentially in real-time. We may identify certain kinds of request, or certain groups of customers, tend to speak in more or less complex manners, and allocate the requests to the appropriate agents. For example, by identifying that those having extremely high complexity dialogs with the automatic agents are inclined to ``break the system'', one could direct this group of users to human agents in the future. 

Moreover, insights gained from analyzing historical datasets can be applied to new service contexts or individuals. That is, one can run the dialog complexity measure on a new dataset and infer characteristics associated with the provided complexity profile. While our experiments served as an illustration of this approach, future research could explore identifying a more complete set of mapping relations between dialog complexity profiles and various contextual, procedural and individual features in service dialogs.

\subsection{Improving Service Agents Evaluation}

A second usage of complexity measure is to improve the evaluation method of service agents. The notion is that, by taking complexity into consideration, agents should be rewarded for handing a more complex dialog with equally satisfying outcome. Here we propose an agent evaluation method that considers dialog complexity and demonstrate its difference with a simulated example. 

Suppose we can have customer support center with $M$ agents. An agent $a_j$ handles  $N_{a_j}$ dialogs in time $T_{a_j}$. A function $\phi(d_i)$ is given to  
find the customer's satisfaction (C-SAT) with a dialog $d_i$ and its complexity is measured by function $c(d_i)$.
The goal is to assess the performance of the agent, represented as $\omega(a_j)$. 

A most basic version of evaluation method is by the average C-SAT ratings  an agent receives, denoted by $\omega_1$ (Equation 6). 

\begin{equation}
 \omega_1(a_j) = \frac{1}{N_{a_j}} * (\sum_{i=1}^{N_{a_j}} \phi(d_i))
\end{equation}

An improved version will take the varying time spent for each service session into consideration, and calculate a weighted sum of C-SAT by the percentage of time (over total time $T_{a_j}$) spent on the corresponding dialog, denoted by $\omega_2$ (Equation 7). 
But the above metrics fail to account for the complexity, i.e., difficulty in handling, of each interaction. 

\begin{equation}
 \omega_2(a_j) = \frac{1}{T_{a_j}} * (\sum_{i=1}^{N_{a_j}} \phi(d_i) * t_i)
\end{equation}

We propose $\omega_3(a_j)$ as defined by Equation 8.
Here, the customer rating of an interaction $i$ is weighted with its complexity $d_i$ and duration $t_i$, and averaged over the whole duration that an agent has to be evaluated. The result is a number which will be between 0-1 if $c$ and C-SAT are in that range.
This metric would also allow agents who work over different time periods ($T_i$) and nature of dialogs to be compared.

\begin{equation}
 \omega_3(a_j) = \frac{1}{T_{a_j}} * (\sum_{i=1}^{N_{a_j}} c(d_i) * \phi(d_i) * t_i)
\end{equation}

To see whether these measures actually make an impact, we consider a simulated scenario using 1000 real dialogs randomly selected from each of the four datasets. We assume that there are 3 agents ($a_1, a_2, a_3$) who handle 300,350 and 350 dialogs in that order.  We assume that the agents cover these dialogs in 30, 40 and 50 hours respectively. Within a time duration $T_{a_i}$, we assume that the agent takes time to handle a dialog proportional to the number of words in it. We assume that each agent is equally trained and were able to achieve a constant C-SAT,$\phi(d_i)$, for any interaction.

\begin{table}
\begin{center}
{\scriptsize
\begin{tabular}{|l|p{1.5cm}|p{1.5cm}|p{1.5cm}|}\hline
&{$\omega_3(a_1)$}&{$\omega_3(a_2)$}&{$\omega_3(a_3)$}
  \\\hline\hline
\multicolumn{4}{|c|}{Random allocation to agents}  \\\hline
Ubuntu&0.450&0.444&0.454\\\hline
Insurance&0.894&0.894& 0.896\\\hline
HR &0.439&0.428&0.453\\\hline
Restaurant &0.542&0.536&0.537\\\hline

\multicolumn{4}{|c|}{Allocation by ascending dialog complexity}  \\\hline
Ubuntu&0.370&0.403&0.483\\\hline
Insurance&0.873&0.894& 0.916\\\hline
HR &0.378&0.425&0.496\\\hline
Restaurant &0.460&0.502&0.601\\\hline
\end{tabular}
}
\caption{Results of simulated experiment to distinguish agents with dialog complexity}\label{tab:eval}
\end{center}
\end{table}

Table~\ref{tab:eval} shows the result of evaluation results of the three agents with the metric we proposed. We consider two 
cases: (1) where the dialogs are assigned to agents randomly and (2) where the allocation is by increasing order of dialog complexity. We see there is sharp difference in measured performance in the second case where agents were given dialogs with different complexity. Our proposed method $\omega_3$ would capture this biased allocation and reward the agent that handled more complex dialogs with equal user satisfaction. On the other hand, conventional metrics such as $\omega_1$ and $\omega_2$ would not have shown any difference.



%% file: conclusion.tex

\mvp
\section{Conclusion}
\label{sec:conclusion}
\mvp
In this paper, we introduced the notion of dialog complexity to understand and compare a collection of dialogs that are routinely 
used in services industry,  proposed a method to calculate it
 and used it to understand customer interactions in a variety of domains at utterance, turn and dialog levels. 
 A dialog complexity measure can conceivably help improving service operation and we discuss its usage for tailoring service handling for varying customer interactions, and demonstrate its usage for improving service evaluation by taking into consideration the difficulty of dialogs that agents handle. 
 
Looking forward, one can extend the current work in many ways. One can  explore deeper dialog content (e.g., n-grams) and structure information, or develop machine learning based approach providing that complexity annotation is available, to create more sophisticated metrics and evaluate whether they can effectively predict the complexity of service dialog handling.  One can also explore using the complexity metric to manage many aspects of service center operation, such as determining the most cost-effective way of handling requests, or even optimizing a contact center dynamically.

